\documentclass[12pt,journal,compsoc,onecolumn]{IEEEtran}

 \usepackage[T1]{fontenc}

\usepackage{lmodern}
\usepackage{graphicx}
\usepackage{microtype}
\usepackage{subfigure}
\usepackage{booktabs,pifont,multirow,multicol} 
\usepackage{amsmath}
\usepackage{amssymb}
\usepackage{mathtools}
\usepackage{pifont}
\usepackage[noadjust]{cite}
\usepackage{caption}

\usepackage{multirow}
\newcommand{\cmark}{\ding{51}}
\newcommand{\xmark}{\ding{55}}

\usepackage[textsize=tiny]{todonotes}

\def\UU{{\bf U}}

\def\V{{\mathbf V}} 
\def\A{{\mathbf \Lambda}}

\def\A{{\bf A}}

\def\W{{\bf W}}
\def\WW{{\bf W}}
\def\SS{{\cal S}}

\def\N{{\cal N}}
\def\G{{\cal G}}
\def\V{{\cal V}}
\def\E{{\cal E}}
\def\F{{\cal F}}

\def\M{{\bf M}}

\def\P{{\bf P}}

\usepackage[ruled,vlined]{algorithm2e}

\title{Learning Coarse-to-Fine Pruning of Graph Convolutional Networks for Skeleton-based Recognition}

\author{Hichem Sahbi
\vspace{1cm}
\\ 
{Sorbonne University, CNRS, LIP6, F-75005, Paris, France}}

\begin{document}
 \maketitle
\begin{abstract}
Magnitude Pruning is a staple lightweight network design method which seeks to remove connections with the smallest magnitude. This process is either achieved in a structured or unstructured manner.  While structured pruning allows reaching high efficiency, unstructured one is more flexible and leads to better accuracy, but this is achieved at the expense of low computational performance. \\
In this paper, we devise a novel coarse-to-fine (CTF) method that gathers the advantages of structured and unstructured pruning while discarding their inconveniences to some extent.  Our method relies on a novel CTF parametrization that models the mask of each connection as the Hadamard product involving four parametrizations which capture channel-wise, column-wise, row-wise and entry-wise pruning respectively. Hence, fine-grained pruning is enabled only when the coarse-grained one is disabled, and this leads to highly efficient networks while being effective. Extensive experiments conducted on the challenging task of skeleton-based recognition, using the standard SBU and FPHA datasets, show the clear advantage of our CTF approach against different baselines as well as the related work.  \\

\noindent {\bf keywords.} {

Coarse and fine-grained pruning \and  graph convolutional networks  \and  skeleton-based recognition.}
\end{abstract}

\section{Introduction}
  
Deep learning (DL) is a rapidly growing subfield of artificial intelligence (AI) which  has made a significant advancement~\cite{Krizhevsky2012} in various pattern recognition tasks including action and hand-gesture classification~\cite{Zhua2016}.  Major actors  in the realm of AI   are nowadays  deploying DL techniques (and particularly neural networks)  to solve problems and gain a competitive edge.   However,   DL's  progress has been achieved  at the expense of a significant increase of time and memory demand,  making it  overpowering to deploy on cheap devices endowed with limited hardware resources.    In the  field of skeleton-based recognition,  graph convolutional networks  (GCNs) are peculiar neural networks that operate on non-euclidean domains (such as skeleton graphs) by  learning  relationships between nodes and edges.   Two categories of GCNs exist in the literature: spectral \cite{kipf17,Li2018} and spatial \cite{Gori2005,attention2019}.    The former relies on the Fourier transform  while  the latter leverages  message passing and multi-head attention  layers.  These layers extract  node representations by aggregating  features  over  their most salient  neighboring nodes,  prior to applying convolutions (as  inner products) on the resulting node aggregates.   With multi-head attention,  spatial GCNs are deemed highly  accurate on skeleton data,  but oversized and computationally overwhelming,  and their deployment on cheap devices requires designing their lightweight counterparts.   \\
\indent  Existing work that addresses the issue of lightweight neural network design includes tensor decomposition~\cite{howard2019}, quantization~\cite{DBLP:journals/corr/HanMD15}, distillation~\cite{DBLP:conf/aaai/MirzadehFLLMG20}, neural architecture search \cite{nasprun} and pruning~\cite{DBLP:conf/nips/CunDS89,DBLP:conf/nips/HassibiS92,DBLP:conf/nips/HanPTD15}. Pruning techniques are particularly effective and consist in  removing unnecessary connections  leading to more compact and faster networks with a minimal decay on accuracy.   One of the mainstream lightweight design  methods  is  magnitude pruning (MP) \cite{DBLP:journals/corr/HanMD15}.   The latter aims  at  ranking network connections according to  the  magnitude, {\it as a proxy to the importance}, of their weights,   prior to  remove the smallest  magnitude connections,  and this eventually leads to  a minimal  impact on performances.   Two categories of MP techniques are widely known: structured~\cite{DBLP:conf/iclr/0022KDSG17,DBLP:conf/iccv/LiuLSHYZ17}  and unstructured ~\cite{DBLP:conf/nips/HanPTD15,DBLP:journals/corr/HanMD15}.   Structured MP consists in removing entire groups of weights, filters or neurons which significantly changes the model architecture and leads to higher compression rates and efficient computation on standard deep learning frameworks/hardware.  However, structured MP suffers from a {\it coarse} pruning granularity as it cannot  target individual (possibly important)  weights within a group,   and this may potentially result into  a significant drop in accuracy particularly when aggressive pruning is achieved.  On another hand,   unstructured MP offers a {\it  fine}  control over granularity as it  identifies and removes  connections individually,   and  maintains the overall network architecture.  Hence,  it  may potentially preserve important connections and   ultimately achieve higher  accuracy compared to structured MP.   However,   unstructured MP  suffers from several downsides including lower compression rate compared to structured MP,  and  slower inference with {\it spread weight distributions} which may be inefficient for acceleration with most of the existing standard hardware.  \\
 \indent In order to gather the upsides of both structured and unstructured pruning techniques,  while mitigating  their downsides,  we devise in this paper a  new pruning method for lightweight GCNs.     The  design principle of our approach  is {\it coarse-to-fine} (CTF) and  achieved using a novel  multi-structured tensor  parametrization; as we traverse this parametrization,  pruning is getting relatively  less structured and computationally less efficient,  but  more  resolute (finer), allowing to reach the targeted pruning rate with a high accuracy.   Given an unpruned network,  we define our  parametrization as the combination of three functions: (i) a band-stop parametrization which keeps only connections with the highest magnitudes,  (ii) a weight-sharing parametrization that groups connections either channel-wise,  column-wise, row-wise or keeps them as singletons,   and (iii) a gating mechanism  which  either keeps weights as singletons,  or removes them in a structured manner.  Besides being able to handle coarse as well as fine-grained pruning,   our composed parametrization allow reaching  a  tradeoff between efficient computation and high accuracy as corroborated through extensive experiments conducted on the challenging task of skeleton-based action and hand-gesture  recognition.  
\section{Related work}

The following review discusses the related work in  pruning and skeleton-based recognition, highlighting the limitations that motivate our contributions.\\

\noindent {\bf Variational Pruning.}  The latter  seeks to learn weights and binary masks in order to capture the topology of pruned networks. This is obtained by minimizing a global loss which mixes  a classification error and a regularizer that controls the cost of the pruned networks \cite{DBLP:conf/iccv/LiuLSHYZ17,REFWen,REFICLR}.  However, existing methods are powerless to implement targeted pruning rates without overtrying multiple weighting of regularizers. Alternative methods explicitly use  $\ell_0$-based criteria to minimize the discrepancy between observed and targeted costs\cite{REFICLR,REFDrop}.   Existing solutions rely on sampling heuristics or relaxation, which promote sparsity  (via  different regularizers including $\ell_1$/$\ell_2$-based,  entropy,  etc.)  \cite{REFGordon,REFCarreira,refref74,refref75}  but are powerless to implement target costs exactly, and may lead to overpruned and thereby disconnected networks. Besides, most of the existing solutions including magnitude pruning  are either structured ~\cite{DBLP:conf/iclr/0022KDSG17,DBLP:conf/iccv/LiuLSHYZ17}  or unstructured ~\cite{DBLP:conf/nips/HanPTD15,DBLP:journals/corr/HanMD15},  and their benefit is not fully explored.  This paper aims to gather the advantages of both structured and unstructured pruning while discarding their limitations.\\

 \noindent {\bf Skeleton-based recognition.}   This task has gained increasing interest due to the emergence of sensors like Intel RealSense   and Microsoft Kinect.  Early methods for hand-gesture and action recognition used RGB~\cite{refref18}, depth~\cite{refref39},  shape/normals~\cite{refref40,refref41,Yun2012,Ji2014,Li2015a,refref59}, and skeleton-based techniques \cite{Wang2018c}. These methods were based on modeling human motions using handcrafted features \cite{Yang2014}, dynamic time warping~\cite{Vemulapalli2014},  temporal information \cite{refref61,refref11}, and temporal pyramids~\cite{Zhua2016}. However, with the resurgence of deep learning, these methods have been quickly overtaken by 2D/3D Convolutional Neural Networks (CNNs)~\cite{refref10,REF3}, Recurrent Neural Networks (RNNs) \cite{Zhua2016,Du2015,Liu2016,DeepGRU,Zhang2017,GCALSTM}, manifold learning~\cite{Huangcc2017,ref23,Liu2021,RiemannianManifoldTraject},  attention-based networks \cite{Song2017},  and GCNs \cite{Lib2018,Yanc2018,Wen2019,Jiang2020}.  The recent emergence of GCNs, in particular,  has led to their increased use in skeleton-based recognition \cite{Li2018}. These models capture spatial and temporal attention among skeleton-joints with better interpretability.  However, when tasks involve relatively large input graphs, GCNs with multi-head attention become computationally inefficient and require lightweight design techniques.  In this paper, we design efficient GCNs that make skeleton-based recognition highly efficient while also being effective.

\section{Graph convnets at glance}

Considering  $\SS=\{\G_i=(\V_i, \E_i)\}_i$ as  a collection of graphs with $\V_i$, $\E_i$ being respectively the nodes and the edges of $\G_i$,  each graph $\G_i$ (denoted for short as $\G=(\V, \E)$) is empowered with a signal $\{\phi(u) \in \mathbb{R}^s: \ v  \in \V\}$ and  an adjacency matrix $\A$.   Graph convolutional networks (GCNs)  learn a set of $C$ filters $\F$ that define convolution on $n$ nodes of  $\G$  as $(\G \star \F)_\V = f\big(\A \  \UU^\top  \   \W\big)$, here $n=|\V|$,  $^\top$ stands for transpose,  $\UU \in \mathbb{R}^{s\times n}$  is the  graph signal, $\W \in \mathbb{R}^{s \times C}$  is the matrix of convolutional parameters corresponding to the $C$ filters and  $f(.)$ is a nonlinear activation applied entry-wise.  With $(\G \star \F)_\V$,   the input signal $\UU$  is projected using $\A$ providing for each node $v$,  the  aggregate set of its neighbors.  Entries  of $\A$ can either be  handcrafted or learned in $(\G \star \F)_\V$  forming a convolutional block with two layers:    the first layer in $(\G \star \F)_\V$ aggregates signals in $\N(\V)$ (as the sets of neighbors of  nodes in $\V$) by multiplying $\UU$ with $\A$, while the second layer performs convolutions by multiplying the resulting aggregates with the $C$ filters in $\W$.  Learning  multiple adjacency (also referred to as attention) matrices (denoted as $\{\A^k\}_{k=1}^K$)  enable capturing  various contexts and graph topologies when achieving  aggregation and convolution. With multiple adjacency  matrices $\{\A^k\}_k$ (and associated convolutional filter parameters $\{\W^k\}_k$),  $(\G \star \F)_\V$ is updated as $f\big(\sum_{k=1}^K \A^k   \UU^\top     \W^k\big)$,  so stacking multiple aggregation and convolutional layers  makes GCNs  more accurate but heavier.
Our  proposed method,   in this paper,  seeks to make GCNs  lightweight yet effective.

\section{Proposed Method}
 
 Subsequently, we formalize a given GCN as a multi-layered neural network $g_\theta$ with weights defined by $\theta = \left\{\WW^1,\dots, \WW^L \right\}$,  and  $L$ its depth, $\WW^\ell \in \mathbb{R}^{d_{\ell-1} \times d_{\ell}}$ its $\ell^\textrm{th}$ 
layer weight tensor, and $d_\ell$ its dimension.  We define the output of a given layer  $\ell$ as
$ \mathbf{\phi}^{\ell} = f_\ell({\WW^\ell}^\top \  \mathbf{\phi}^{\ell-1})$, $\ell \in \{2,\dots,L\}$,  with $f_\ell$ an activation function; without a loss of generality, we omit the bias in the definition of  $\mathbf{\phi}^{\ell}$.\\
Pruning is the process of removing a subset of weights in $\theta$ by multiplying $\WW^\ell$  with a binary mask $\M^\ell \in \{ 0,1 \}^{d_{\ell-1} \times d_{\ell}}$. The binary entries of $\M^\ell$ are determined by pruning the underlying layer connections,  so $\mathbf{\phi}^{\ell} = f_\ell((\M^\ell \odot \WW^\ell)^\top \ \mathbf{\phi}^{\ell-1} )$,  with $\odot$ being the element-wise matrix product.  In our definition of  pruning, entries of the tensor $\{\M^\ell\}_\ell$ are set depending on the prominence and also on how the underlying connections in $g_\theta$ are grouped (or not); pruning that removes all the connections individually (resp. jointly) is referred to as {\it unstructured} (resp. {\it structured}) whilst pruning that removes some connections {\it first} group-wise and {\it then} individually is dubbed as {\it coarse-to-fine.}  In what follows,  we introduce our main contribution; a novel coarse-to-fine  method that allows combining multiple pruning granularities resulting into efficient and also effective lightweight networks (as shown later in experiments).   

\subsection{Coarse-to-fine Pruning} 

We define our  parametrization as  the Hadamard product involving a weight tensor and a {\it cascaded}  function applied to the same tensor as
\begin{eqnarray}\label{eq2}
  \WW^\ell = \hat{\WW}^\ell \odot \psi (\hat{\WW}^\ell),
\end{eqnarray}
here   $\hat{\WW}^\ell$ is a latent tensor and $\psi(\hat{\WW}^\ell)$ a continuous relaxation of $\M^\ell$ which enforces the prior that (i)  weights $\hat{\WW}^\ell$  with the smallest magnitude should be removed, (ii) entries in  mask  $\psi(\hat{\WW}^\ell)$ are either removed group-wise (through  rows,  columns,  channels) or removed individually.   In the following,  we expand the definition of fine and coarse parametrizations (respectively denoted as $\psi_f$ and $\psi_c$) prior to their combination in Eq.~\ref{eq55556}. Unless stated otherwise,  we omit $\ell$ in the definition of  $\hat{\WW}^\ell$ and we rewrite it (for short) as $\hat{\WW}$.\\ 

\noindent {\bf Fine-grained  parametrization.} As subsequently described, the  function $\psi_f(\hat{\WW})$ is entry-wise applied to the tensor  $\hat{\WW}$ with the prior that small magnitude weights should be individually removed.   The class of $\psi_f$ functions must be:  (1) differentiable,   (2) symmetric,    (3) bounded in $[0,1]$,  and  (4) asymptotically reaching  $1$ when entries of $\psi_f(.)$ have large magnitude and $0$ otherwise.   Properties (1) and (2)  respectively  ensure  that $\psi_f$ has computable gradient  and  that only the magnitude of the latent weights matters whereas properties (3) and (4) guarantee that  $\psi_f$ is neither overflowing  nor  changing the sign of the latent weights,  and also values  in $\psi_f$ behave as crisp (almost binary) masks reaching asymptotically $1$ iff the latent weights in $|\hat{\WW}|$ are sufficiently large,  and $0$ otherwise. In practice, a choice of $\psi_f$ that satisfies these four conditions is the symmetrized shifted  sigmoid $\psi_f(\hat{\WW})=2 \ \textbf{sigmoid}(\sigma \hat{\WW}^2)-1$;  here  power and sigmoid are applied entry-wise and  $\sigma$  is  a scaling factor that controls the crispness (binarization) of mask entries in $\psi_f (\hat{\WW})$.   In practice,  $\sigma$ is annealed so as to cut-off the connections in the network in smooth and differentiable manner --- as the optimization of $ \hat{\WW}$   evolves ---  while obtaining  at the end of the optimization process crisp (almost binary)  masks.  \\  

\noindent {\bf Coarse-grained  parametrization.} The  function $\psi_c(\hat{\WW})$   implements a coarse-grained pruning by removing connections group-wise (row-wise,  column-wise  or block/channels-wise)  in the tensor $\hat{\WW}$.  This function  is formally defined as 

\begin{equation}\label{eq5555}
\begin{array}{lll}
 \psi_c(\hat{\WW})&=&\underbrace{\phi^{-1}(\P_r \  \phi(\hat{\WW}))}_{\textrm{row-wise pruning}} \odot \underbrace{\phi^{-1}({\phi (\hat{\WW})^\top \ \P_c})}_{\textrm{column-wise pruning}} \\
     &  &\ \ \  \ \ \  \ \ \ \ \ \  \ \ \ \ \ \ \ \ \ \ \ \ \odot  \underbrace{\phi^{-1}(\P_{r}\P_c^\top \  \phi(\hat{\WW}))}_{\textrm{block-wise pruning}},
\end{array}
\end{equation} 
here $\phi$  (resp. $\phi^{-1}$)  reshapes a matrix into a vector (resp. vice-versa), and  $\P_r \in \{0,1\}^{(d_{\ell-1}\times d_{\ell})^2}$, $\P_c \in \{0,1\}^{(d_{\ell-1} \times d_\ell)^2}$ are two adjacency matrices that model the neighborhood across respectively  the rows and the columns of $\hat{\WW}$  whilst $\P_{r} \P^\top_{c}\in \{0,1\}^{(d_{\ell-1} \times d_\ell)^2}$ models this neighborhood through blocks/channels of the tensor  $\hat{\WW}$.   \\

\begin{figure}[hpbt]
   \begin{center}
     \centerline{\scalebox{0.99}{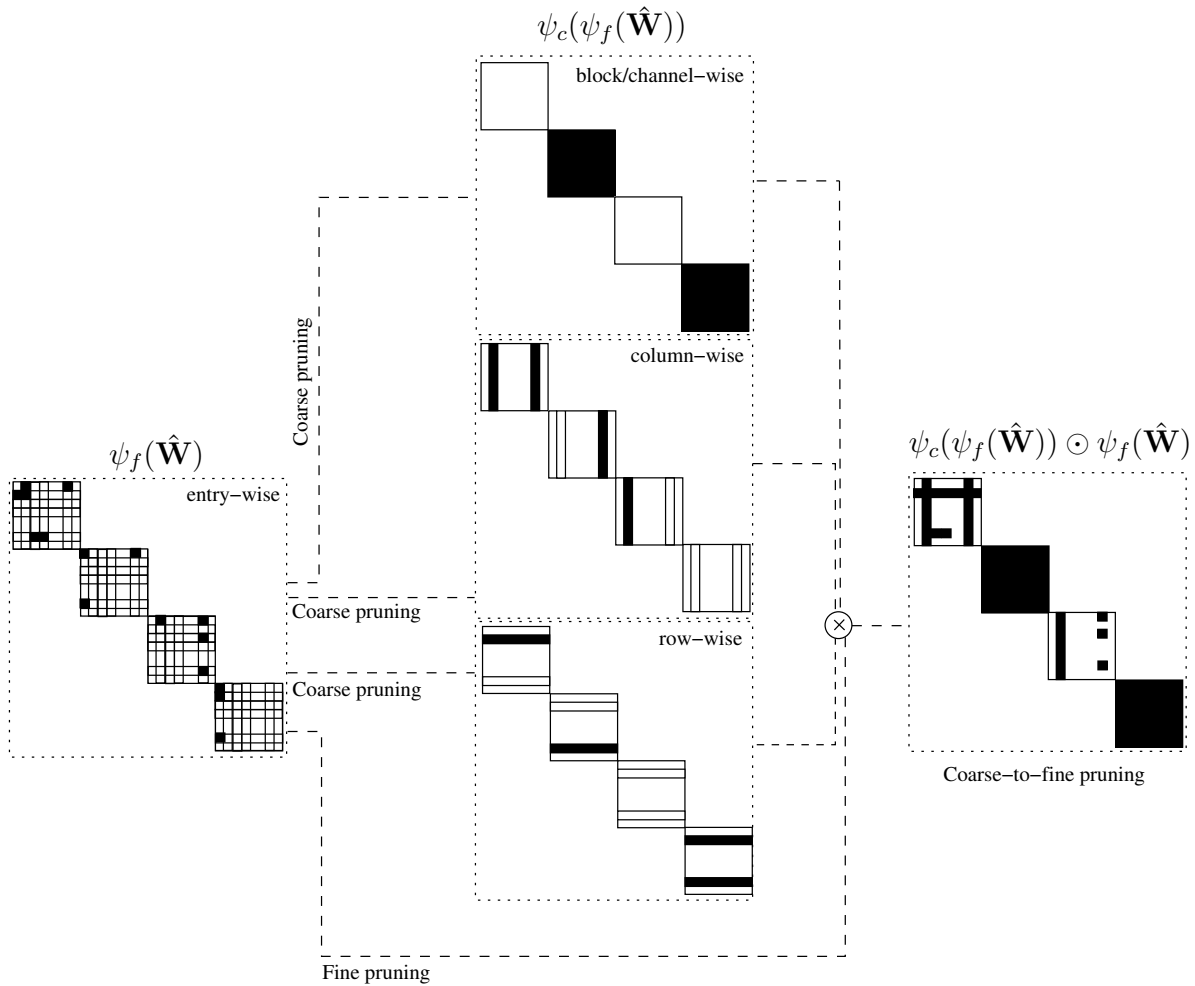}}
     \caption{This figure shows the  CTF pruning process in Eq.~\ref{eq55556}; here each diagonal block corresponds to a channel.} \label{fig123}
 \end{center}
 \end{figure}

\noindent {\bf Coarse-to-fine-grained parametrization.} Considering the above definition of $\psi_c$  and $\psi_f$,   we obtain our complete coarse-to-fine   mask parametrization as 
 \begin{equation}\label{eq55556}
 \psi(\hat{\WW}) =\underbrace{[\psi_c (\psi_f(\hat{\WW}))]}_{\textrm{coarse-grained pruning}} \odot  \underbrace{\psi_f(\hat{\WW})}_{\textrm{fine-grained pruning}}
 \end{equation} 
 From Eqs.~\ref{eq5555} and~\ref{eq55556}, assuming crisp (almost binary)  entries in  $\hat{\WW}$  (thanks to the sigmoid),   block-wise pruning has the highest priority,   followed by column-wise and then row-wise pruning.  This priority allows designing highly efficient lightweight networks with a coarse-granularity for block / column / row-wise (structured) pruning while  the entry-wise (unstructured) parametrization  is less computationally efficient but allows reaching the targeted pruning rate with a finer granularity (see Fig.~\ref{fig123}).  In sum, CTF allows efficient coarse-grained network design while also leveraging the accuracy of fine-grained one,  thereby leading to {\it both} efficient and effective pruned networks as shown subsequently in experiments.

\subsection{Variational Pruning} 
 
By considering Eq.~\ref{eq2}, we define our CTF  pruning loss as
\begin{equation}\label{eq366}
\begin{array}{lll}
{\cal L}_e\big( \{ \psi(\hat{\WW}^{\ell}) \odot \hat{\WW }^{\ell} \}_\ell\big) + \lambda \displaystyle \bigg(\sum_{\ell=1}^{L-1}  \psi(\hat{\WW}^{\ell}) \ - c\bigg)^2,
\end{array}     
\end{equation} 
 here the left-hand side term is the cross entropy loss that measures the discrepancy between predicted and ground-truth labels. The right-hand side term is a budget loss that allows reaching any targeted pruning cost $c$.   In the above objective function, $\lambda$ is overestimated (to 1000 in practice) in order to make Eq.~\ref{eq366} focusing on the implementation of the budget.  As training reaches its final epochs, the budget loss  reaches its minimum and the gradient of the global objective function becomes dominated by the gradient of ${\cal L}_e$, and this allows improving further classification performances.
\section{Experiments}
In this section, we evaluate the performances of our pruned GCNs on skeleton-based recognition using two challenging datasets, namely SBU \cite{Yun2012}  and FPHA \cite{Garcia2018}. SBU is an interaction dataset acquired using the Microsoft Kinect sensor; it includes in total 282 moving skeleton sequences (performed by two interacting individuals) belonging to 8 categories. Each pair of interacting individuals corresponds to two 15 joint skeletons and each joint is characterized with a sequence of its 3D coordinates across video frames.   In this dataset, we consider the same evaluation protocol as the one suggested in the original dataset release \cite{Yun2012}  (i.e., train-test split). The FPHA dataset includes 1175 skeletons belonging to 45 action categories with high inter and intra subject variability.  Each skeleton includes 21 hand joints and each joint is again characterized with a sequence of its 3D coordinates across video frames. We evaluate the performance of our method on FPHA following the protocol in \cite{Garcia2018}.  In all these experiments, we report the average accuracy over all the classes of actions.\\
 \begin{table}
 \centering
\resizebox{0.5\columnwidth}{!}
{
\begin{tabular}{cc|c}
{\bf Method}      &   & {\bf Accuracy (\%)}\\
\hline 
  Raw Position \cite{Yun2012} & $ \ $   & 49.7   \\ 
  Joint feature \cite{Ji2014}  & $ \ $   & 86.9   \\
  CHARM \cite{Li2015a}       & $ \ $    & 86.9   \\
 \hline  
H-RNN \cite{Du2015}         & $ \ $    & 80.4   \\ 
ST-LSTM \cite{Liu2016}      & $ \ $    & 88.6    \\ 
Co-occurrence-LSTM \cite{Zhua2016} & $ \ $  & 90.4  \\ 
STA-LSTM  \cite{Song2017}     & $ \ $   & 91.5  \\ 
ST-LSTM + Trust Gate \cite{Liu2016} & $ \ $  & 93.3 \\
VA-LSTM \cite{Zhang2017}      & $ \ $  & 97.6  \\
 GCA-LSTM \cite{GCALSTM}                    &   $ \ $      &  94.9     \\ 
  \hline
Riemannian manifold. traj~\cite{RiemannianManifoldTraject} &  $ \ $  & 93.7 \\
DeepGRU  \cite{DeepGRU}        &    $ \ $   &    95.7    \\
RHCN + ACSC + STUFE \cite{Jiang2020} & $ \ $   & 98.7 \\ 
  \hline
\hline 
  Our baseline GCN &              &        98.4      
\end{tabular}}
\caption{Comparison of our baseline GCN against related work on the SBU database.}\label{tab222}
\end{table}

\begin{table}
\centering
\resizebox{0.65\columnwidth}{!}{
\begin{tabular}{ccccc}
{\bf Method} & {\bf Color} & {\bf Depth} & {\bf Pose} & { \bf Accuracy (\%)}\\
\hline
  2-stream-color \cite{refref10}   & \cmark  &  \xmark  & \xmark  &  61.56 \\
 2-stream-flow \cite{refref10}     & \cmark  &  \xmark  & \xmark  &  69.91 \\  
 2-stream-all \cite{refref10}      & \cmark  & \xmark   & \xmark  &  75.30 \\
\hline 
HOG2-dep \cite{refref39}        & \xmark  & \cmark   & \xmark  &  59.83 \\    
HOG2-dep+pose \cite{refref39}   & \xmark  & \cmark   & \cmark  &  66.78 \\ 
HON4D \cite{refref40}               & \xmark  & \cmark   & \xmark  &  70.61 \\ 
Novel View \cite{refref41}          & \xmark  & \cmark   & \xmark  &  69.21  \\ 
\hline
1-layer LSTM \cite{Zhua2016}        & \xmark  & \xmark   & \cmark  &  78.73 \\
2-layer LSTM \cite{Zhua2016}        & \xmark  & \xmark   & \cmark  &  80.14 \\ 
\hline 
Moving Pose \cite{refref59}         & \xmark  & \xmark   & \cmark  &  56.34 \\ 
Lie Group \cite{Vemulapalli2014}    & \xmark  & \xmark   & \cmark  &  82.69 \\ 
HBRNN \cite{Du2015}                & \xmark  & \xmark   & \cmark  &  77.40 \\ 
Gram Matrix \cite{refref61}         & \xmark  & \xmark   & \cmark  &  85.39 \\ 
TF    \cite{refref11}               & \xmark  & \xmark   & \cmark  &  80.69 \\  
\hline 
JOULE-color \cite{refref18}         & \cmark  & \xmark   & \xmark  &  66.78 \\ 
JOULE-depth \cite{refref18}         & \xmark  & \cmark   & \xmark  &  60.17 \\ 
JOULE-pose \cite{refref18}         & \xmark  & \xmark   & \cmark  &  74.60 \\ 
JOULE-all \cite{refref18}           & \cmark  & \cmark   & \cmark  &  78.78 \\ 
\hline 
Huang et al. \cite{Huangcc2017}     & \xmark  & \xmark   & \cmark  &  84.35 \\ 
Huang et al. \cite{ref23}           & \xmark  & \xmark   & \cmark  &  77.57 \\  
\hline 
HAN  \cite{Liu2021}   & \xmark  & \xmark   & \cmark & 85.74 \\
  \hline
  \hline
Our  baseline GCN                   & \xmark  & \xmark   & \cmark  &  86.43                                                 
\end{tabular}}
\caption{Comparison of our baseline GCN against related work on the FPHA database.}\label{compare2}
\end{table}

\noindent {\bf Implementation details and  baseline GCNs.}  All the GCNs have been trained using the Adam optimizer for $2,700$ epochs with a batch size of $200$ for SBU and $600$ for FPHA, a momentum of $0.9$, and a global learning rate  (denoted as $\nu(t)$)  inversely proportional to the speed of change of the loss used to train the networks; with $\nu(t)$ decreasing as $\nu(t) \leftarrow \nu(t-1) \times 0.99$ (resp. increasing as $\nu(t) \leftarrow \nu(t-1) \slash 0.99$) when the speed of change of the loss in Eq.~\ref{eq366} increases (resp. decreases).  Experiments were run on a GeForce GTX 1070 GPU device with 8 GB memory, without dropout or data augmentation. The baseline GCN architecture for SBU includes an attention layer of 1 head, a convolutional layer of 8 filters, a dense fully connected layer, and a softmax layer; notice that this architecture is not very heavy, nonetheless its pruning is very challenging (particularly at high pruning rates) as it may result into disconnected networks. The baseline  GCN architecture for FPHA is heavier and includes 16 heads, a convolutional layer of 32 filters, a dense fully connected layer, and a softmax layer.  Both the baseline (unpruned) GCN architectures, on the SBU and the FPHA benchmarks, are accurate  (see tables.~\ref{tab222} and~\ref{compare2}), and our goal is to make them lightweight while maintaining their accuracy.\\
 \begin{table}[h]
 \begin{center}
   \resizebox{0.65\columnwidth}{!}{
  \begin{tabular}{ccll}    
   \rotatebox{0}{Pruning rates}  &     \rotatebox{0}{Accuracy (\%)} & SpeedUp  & \rotatebox{0}{Observation}  \\
 \hline
  \hline
    0\%    &    \bf98.40   & none  & Baseline GCN\\
 
                                70\% &  93.84 & none  &  Band-stop Weight Param.\\

    \hline
  \multirow{3}{*}{\rotatebox{0}{90\%}}     &  83.07    &  11$\times$  & Coarse MP (structured)  \\

                                                              &   96.92  & none  & Fine MP (unstructured)    \\   

                                                              &   89.23    & 6$\times$ &   Coarse-to-Fine MP (both)             \\   
    \hline
  \multirow{3}{*}{\rotatebox{0}{95\%}}     &75.38  &  34$\times$  & Coarse MP (structured)  \\
                                                              &   93.84   & none   & Fine MP (unstructured)    \\   

                                                              &    84.61  & 9$\times$  &  Coarse-to-Fine MP (both)             \\

\hline 
  \multirow{3}{*}{\rotatebox{0}{98\%}}     & 49.23  &  235$\times$  & Coarse MP (structured)  \\

                                                              &   90.76 & none  &  Fine MP (unstructured)    \\   

                                                              &   76.92   &  43$\times$    & Coarse-to-Fine MP (both)             \\

    \hline \hline
  \multicolumn{4}{c}{Comparative (regularization-based) pruning}   \\                              
    \hline 
      \multirow{4}{*}{\rotatebox{0}{98\%}}                                &    55.38 & none & MP+$\ell_0$-reg. \\
                                                         &    73.84 & none  & MP+$\ell_1$-reg. \\                                                                                                                                                                      
                                 &    61.53 & none &  MP+Entropy-reg. \\ 
                                &   75.38 & none & MP+Cost-aware-reg.

  \end{tabular}}
\end{center}
\caption{This table shows detailed performances and ablation study on SBU for different  pruning rates. ``none'' stands for no-actual speedup is observed as the underlying tensors/architecture remain  shaped identically to the unpruned network (despite having pruned connections); see also Fig.~\ref{tab21fig}.}\label{table21}
\end{table}

 \begin{table}[h]
 \begin{center}
\resizebox{0.69\columnwidth}{!}{
  \begin{tabular}{ccll}    
   \rotatebox{0}{Pruning rates}  &     \rotatebox{0}{Accuracy (\%)} & SpeedUp & \rotatebox{0}{Observation}  \\
 \hline
  \hline
    0\%    &    \bf86.43   & none  & Baseline GCN\\
 
                                50\% & 85.56 &  none & Band-stop Weight Param.\\

    \hline
  \multirow{3}{*}{\rotatebox{0}{90\%}}     & 76.69  &  13$\times$ & Coarse MP (structured)  \\
                                                              &   83.13   & none  & Fine MP (unstructured)    \\   
                                                              &   80.17   &  6$\times$ & Coarse-to-Fine MP (both)             \\   
    \hline
  \multirow{3}{*}{\rotatebox{0}{95\%}}     & 70.08  & 37$\times$ & Coarse MP (structured)  \\
                                                              &   81.56   &  none & Fine MP (unstructured)    \\   
                                                              &   77.56   & 13$\times$  & Coarse-to-Fine MP (both)             \\

\hline 
  \multirow{3}{*}{\rotatebox{0}{98\%}}     &  63.30  &  96$\times$ & Coarse MP (structured)  \\

                                                              &   76.86   &  none & Fine MP (unstructured)    \\   

                                                              &   70.95   & 41$\times$  & Coarse-to-Fine MP (both)             \\   

    \hline \hline
  \multicolumn{4}{c}{Comparative (regularization-based) pruning}   \\                              
    \hline 
      \multirow{4}{*}{\rotatebox{0}{98\%}}                                      &    64.69 & none  & MP+$\ell_0$-reg. \\
                                                         &   70.78 & none & MP+$\ell_1$-reg. \\                                                                                                                                                                      
                                 &    67.47 & none & MP+Entropy-reg. \\ 
                                &   69.91 & none & MP+Cost-aware-reg.

  \end{tabular}}
\end{center}
\caption{This table shows detailed performances and ablation study on FPHA for different  pruning rates. ``none'' stands for no-actual speedup is observed as the underlying tensors/architecture remain shaped identically to the unpruned network (despite having pruned connections); see also Fig.~\ref{tab21fig}.}\label{table22}
\end{table}
 \begin{figure*}
\centering

\includegraphics[width=0.33\linewidth]{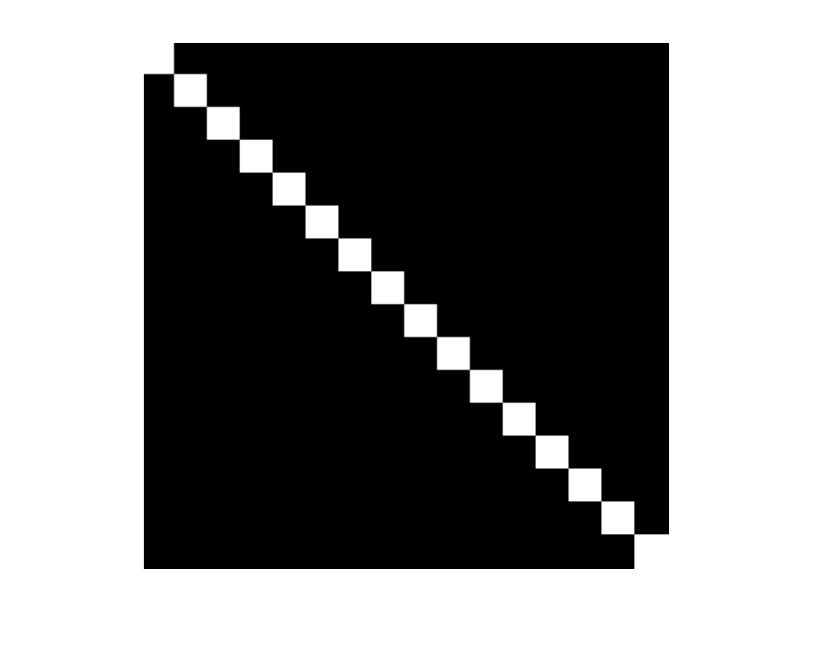}\\
\resizebox{1.\textwidth}{!}{
  \includegraphics[width=1.5\linewidth]{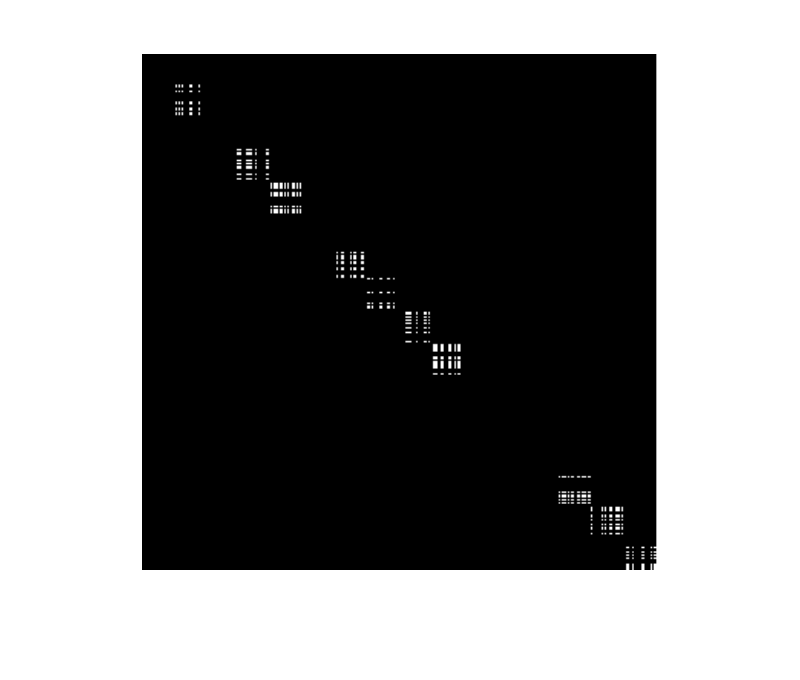} \includegraphics[width=1.5\linewidth]{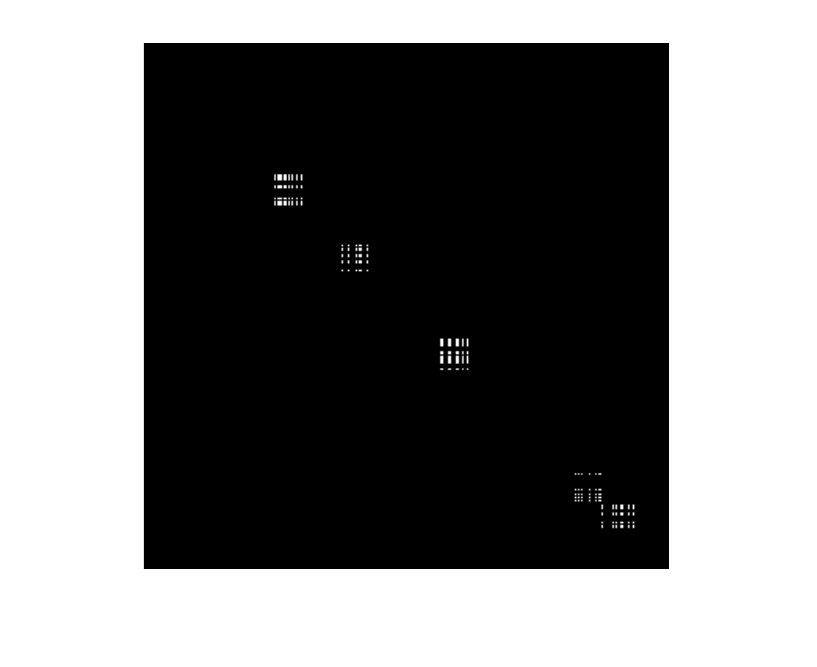} \includegraphics[width=1.5\linewidth]{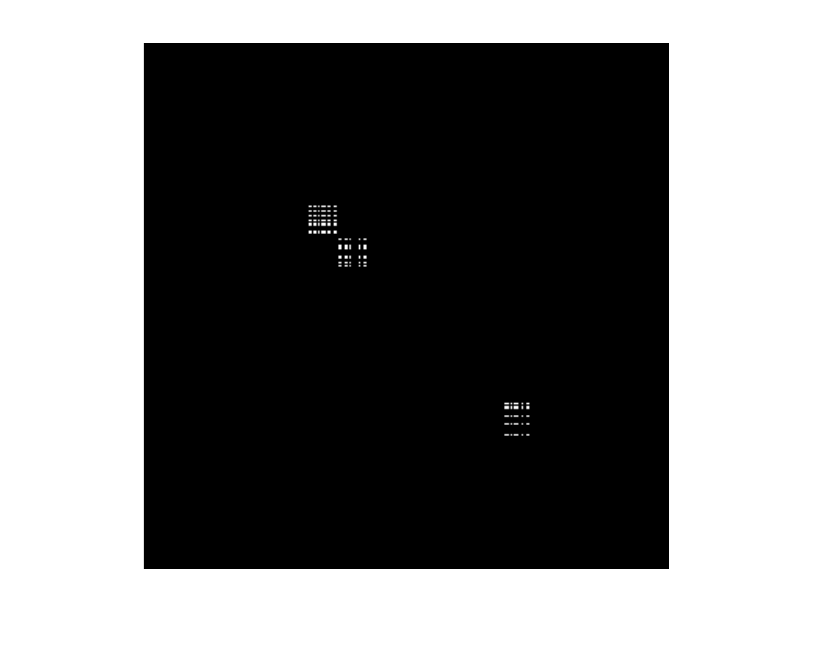}} \\
\resizebox{1.\textwidth}{!}{
  \includegraphics[width=1.5\linewidth]{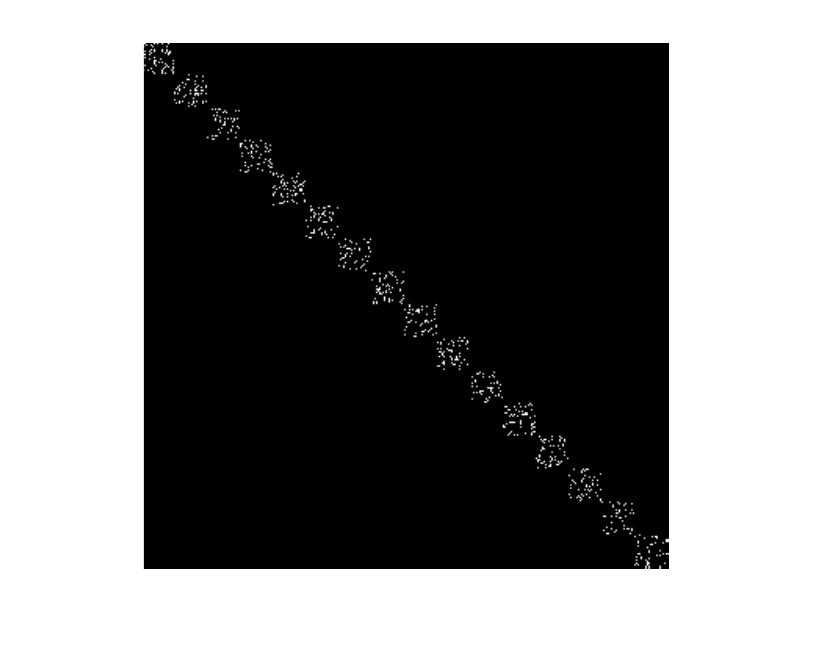} \includegraphics[width=1.5\linewidth]{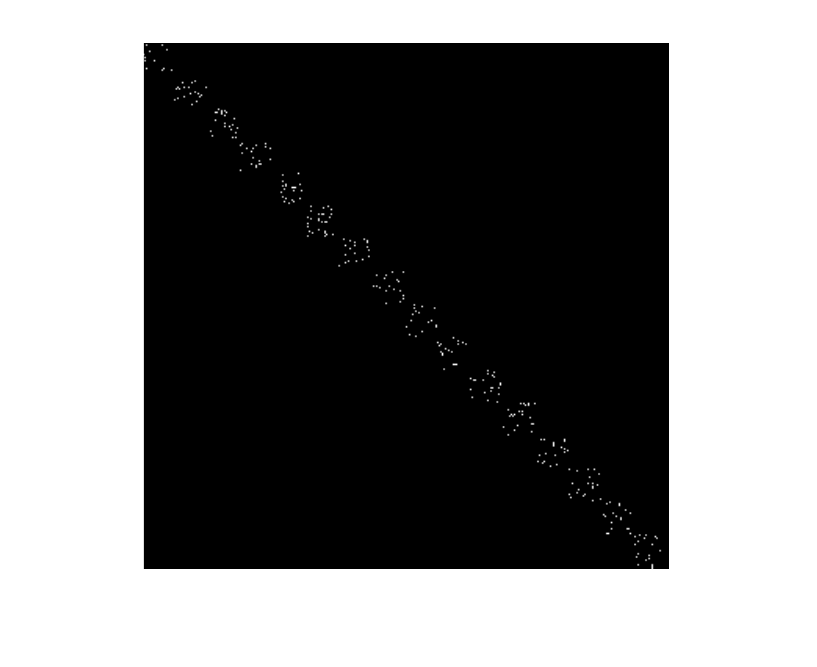} \includegraphics[width=1.5\linewidth]{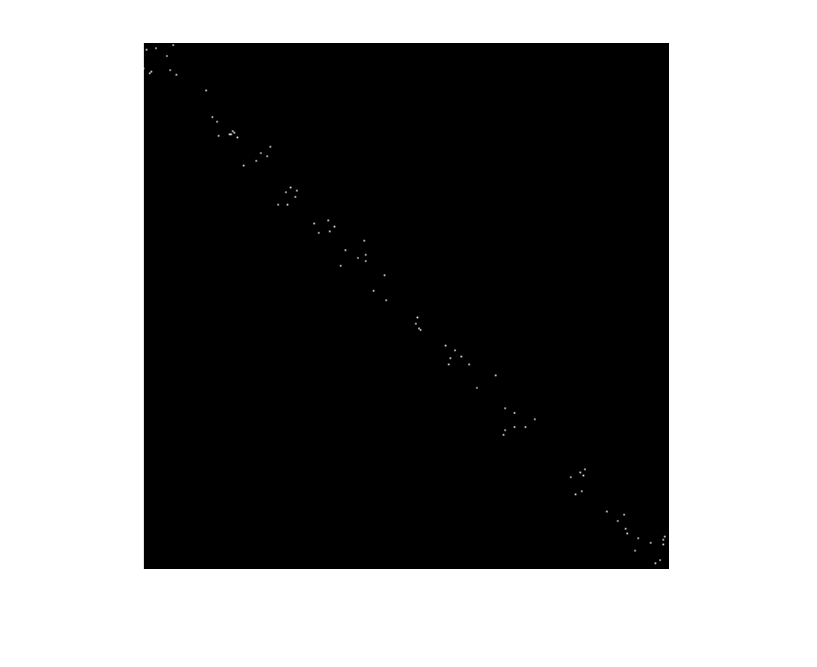}} \\
\resizebox{1.\textwidth}{!}{
  \includegraphics[width=1.5\linewidth]{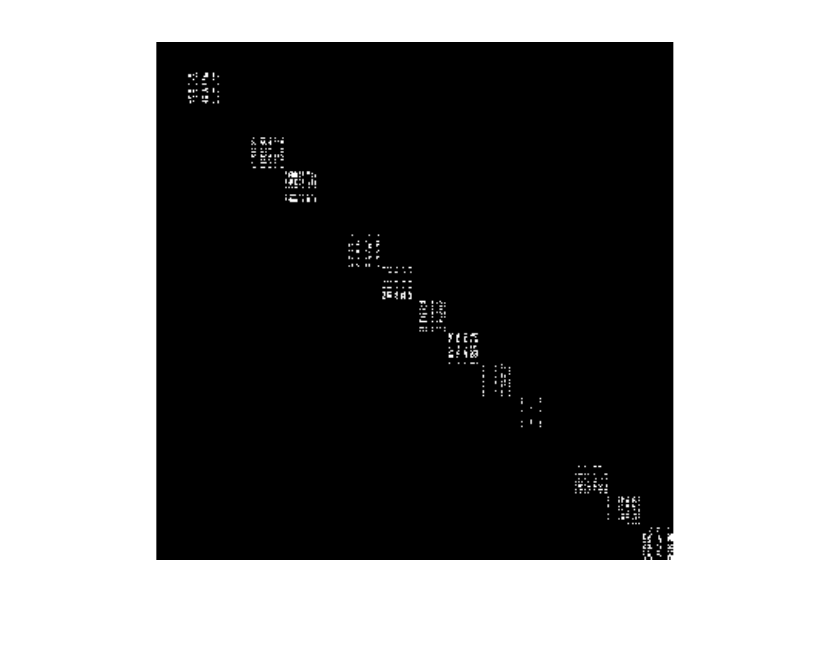} \includegraphics[width=1.5\linewidth]{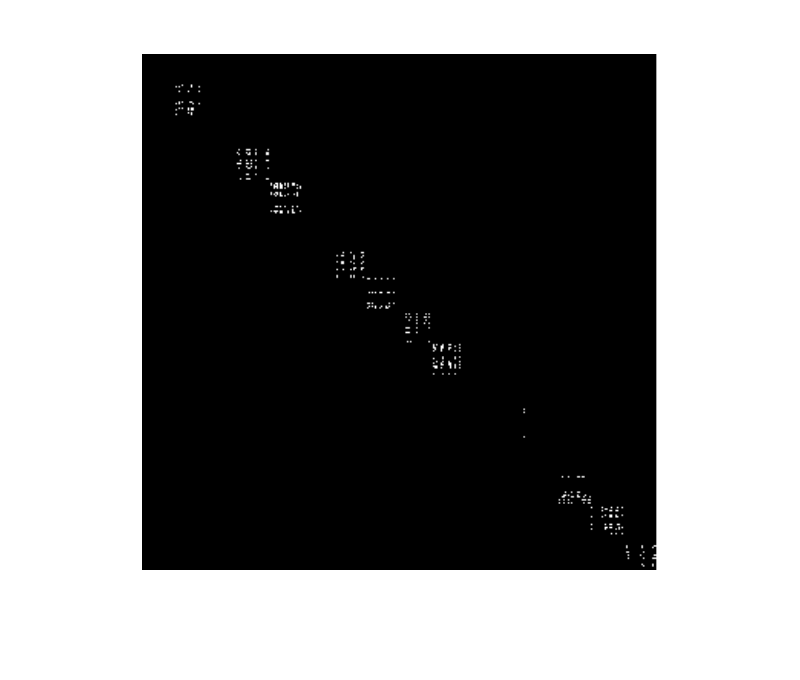} \includegraphics[width=1.5\linewidth]{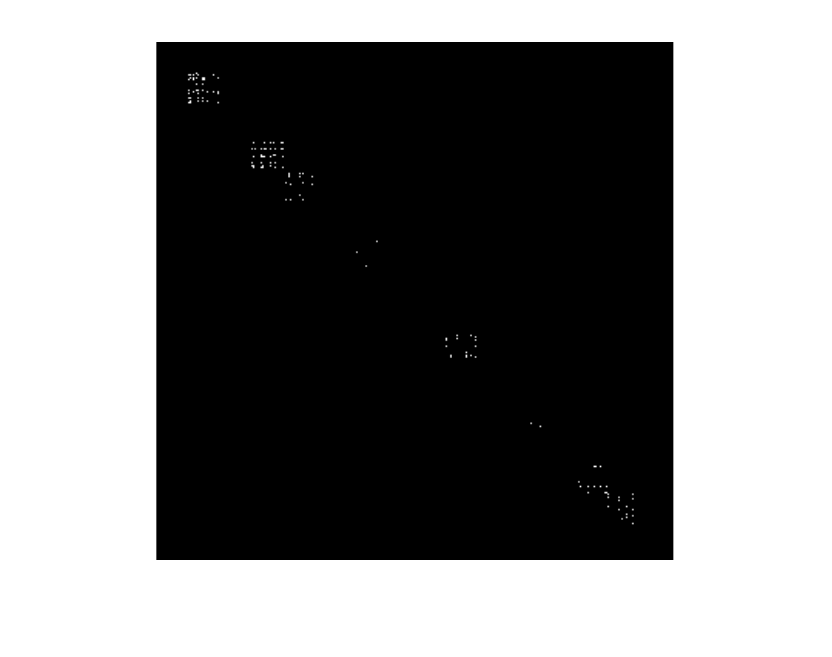}}
\caption{This figure shows a crop of the mask tensor of the second (multi-head-attention) layer of our GCNs when trained on the FPHA dataset. Top row corresponds to the original mask (without pruning) while the second and the third rows correspond to masks obtained with structured and unstructured pruning respectively (with increasing pruning rates; from left-to-right equal to 90\%, 95\% and 98\% respectively). The final row corresponds to masks obtained with semi-structured pruning (with again increasing pruning rates; from left-to-right, equal to 90\%, 95\% and 98\% respectively). In all these masks, each diagonal block corresponds to a channel. Better to zoom the PDF.}\label{tab21fig}
\end{figure*}

\noindent {\bf Lightweight GCNs (Comparison and Ablation).} Tables~\ref{table21}-\ref{table22}  show a comparison and an ablation study of our method both on SBU and FPHA datasets.  First,  according to to tables \ref{table21}-\ref{table22},  when only the cross entropy loss is used without budget   (i.e., $\lambda=0$ in  Eq. \ref{eq366}), performances are close  to the initial heavy GCNs (particularly on FPHA), with less parameters\footnote{Pruning rate does not exceed 70\% and no control on this rate is achievable.} as this produces a regularization effect similar to \cite{dropconnect2013}. Then,   when pruning is achieved with the coarse-grained parametrization,  the accuracy is relatively low but the speedup is high particularly for high pruning regimes.  When pruning is performed with the fine-grained parametrization, the accuracy reaches its highest value,  but no speedup is observed as pruning is unstructured and the architecture of the pruned networks remains unchanged.   When the coarse-to-fine parametrization is used,  we observe the best tradeoff between accuracy and speedup;  in other words,  coarsely pruned parts of the network lead to high speedup and efficient computation,  while finely pruned parts allow reaching better accuracy with a limited impact on computation,  so a significant speedup is still observed.  \\
\indent Extra comparison against other regularizers shows the substantial gain of our method.  Indeed, our method is compared against different variational pruning with regularizers  plugged in Eq. \ref{eq366} instead of our budget loss,  namely $\ell_0$ \cite{REFICLR},  $\ell_1$ \cite{refref74},  entropy \cite{refref75} and $\ell_2$-based cost  \cite{REFLemaire}; all without parametrization.   From the observed results,   the impact of  our method  is substantial for different settings  and for  equivalent pruning rate (namely 98\%).  Note that when alternative regularizers are used, multiple settings (trials) of the underlying hyperparameter $\lambda$ (in Eq. \ref{eq366}) are considered prior to reach the  targeted rate, and this makes the whole training and pruning process overwhelming.  While cost-aware regularization makes training more tractable, its downside resides in the observed collapse of trained masks; this is a well known effect that affects performances  at high pruning rates.   Finally,  Fig.\ref{tab21fig} shows examples of obtained mask tensors taken from the second (attention) layer of the pruned GCNs; we observe compact tensor weight distributions with some individually pruned connections when using  CTF,   while coarse-grained and fine-grained pruning, when taken  separately, either produce {\it spread} or {\it compact} tensors with  a negative impact on either {\it speed} or {\it accuracy} respectively.  CTF gathers both fine  and coarse-grained  advantages while discarding   their downsides. \\

\section{Conclusion} 
We introduce in this paper a CTF approach for pruning.  The strength of the proposed method resides in its ability to combine the advantages of coarse-grained (structured) and fine-grained (unstructured) pruning while discarding their downsides.  The proposed method relies on a novel weight parametrization that first prune tensors channel-wise, column-wise, then row-wise, and finally entry-wise  enabling both efficiency (with a fine implementation of pruning, low-rank tensors) and high accuracy.  Experiments conducted on the challenging tasks of action and hand-gesture recognition, using two standard datasets, corroborate these findings.

{

}

\end{document}